# SynGen-Vision: Synthetic Data Generation for training industrial vision models

Alpana Dubey
Accenture Labs
Bangalore, India
alpana.a.dubey@accenture.com

Suma Mani Kuriakose
Accenture Labs
Bangalore, India
suma.mani.kuriakose@accenture.com

Nitish Bhardwaj
Accenture Labs
Bangalore, India
nitish.a.bhardwaj@accenture.com

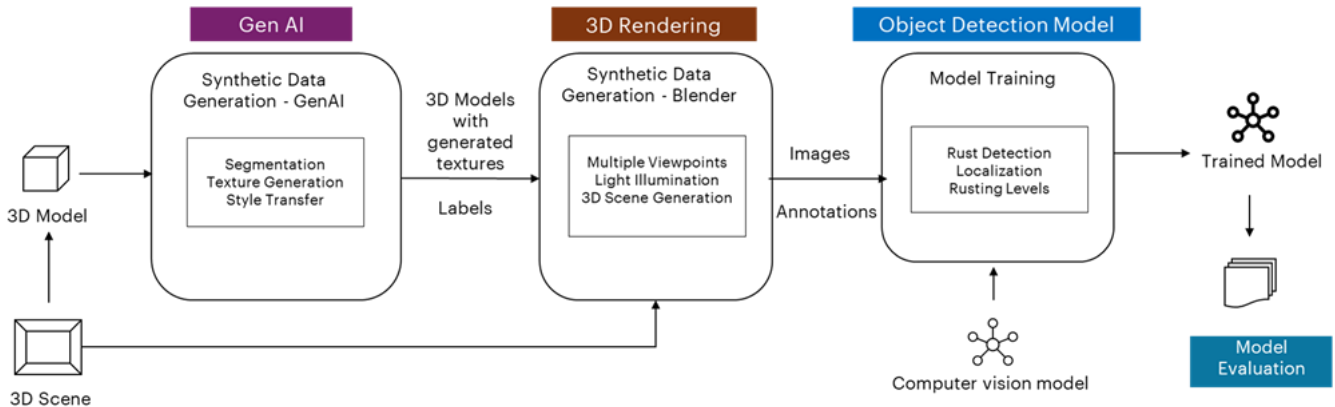

Figure 1: SynGen-Vision: End-to-end pipeline for Synthetic Data Generation using GenAI, and 3D modeling to train a computer vision model for Rust Detection

## Abstract

We propose an approach to generate synthetic data to train computer vision (CV) models for industrial wear and tear detection. Wear and tear detection is an important CV problem for predictive maintenance tasks in any industry. However, data curation for training such models is expensive and time-consuming due to the unavailability of datasets for different wear and tear scenarios. Our approach employs a vision language model along with a 3D simulation and rendering engine to generate synthetic data for varying rust conditions. We evaluate our approach by training a CV model for rust detection using the generated dataset and tested the trained model on real images of rusted industrial objects. The model trained with the synthetic data generated by our approach, outperforms the other approaches with a mAP50 score of 0.87. The approach is customizable and can be easily extended to other industrial wear and tear detection scenarios.

## 1 Introduction

The availability of a high-quality and representative dataset is crucial for the performance of computer vision (CV) algorithms. However, curation of such a dataset is time consuming and effort intensive as it requires access to data and requires human effort to label the data. Moreover, in many situations, data is extremely scarce and non-existent. For example, it is practically impossible to have data of an industrial equipment with different levels of rust until they are deployed and used for some time. Synthetic data generation approaches try to address such challenges by computationally generating representative datasets for a variety of CV tasks [11, 18].

Synthetic data generators mimic real world data by generating accurately labeled, photorealistic training images from all angles, incorporating various lighting conditions. This generated data helps in drastically lowering costs and increasing the performance of CV detection models.

Most of the existing synthetic data generation approaches aim at generating labeled data for object detection and classification tasks [11, 18]. Therefore, they are often not suitable for CV tasks where the objective is to detect other attributes in the images; for example, detection of rust, cracks, aging of equipment, etc. To train a CV model for detecting industrial wear and tear, one needs to simulate such wear and tear under varying environmental conditions and occlusion levels mimicking the real world [6, 14].

In this paper, we propose a synthetic data generation approach aimed at training computer vision models for detecting industrial wear and tear (as shown in Figure 1). More specifically, we demonstrate our approach on rust detection. Our approach simulates an industrial environment in a 3D space using a 3D simulation and rendering engine. The approach takes user prompts describing the wear and tear conditions of interest as input. The overall approach consists of three steps. First, the user prompt is automatically refined and used to generate several variations in the environment and texture maps to simulate the specified wear and tear conditions. Second, the generated texture maps are applied to 3D models. Third, synthetic data is generated from varying angles, lighting conditions, and distances by the rendering engine. We compare our approach with other prevailing Generative AI (GenAI) approaches for synthetic data generation and achieve an accuracy of 0.87.

## 2 Related work

We present here the related work along three broad topics: first, around industrial wear and tear detection, second around synthetic data generation and third around the use of GenAI for industrial use-cases.

Industrial wear and tear detection is an active area of research considering cost and optimization factors. Existing approaches of data collection for industrial wear and tear involve hyperspectral imaging for industrial applications [13, 15, 20]. Most of the deep neural based solutions for wear and tear detection find data gathering and manual annotation very challenging, expensive and time consuming [1, 6, 13, 14, 20]. Our approach addresses this limitation using synthetic data generation.

The other body of work around synthetic data generation includes 3D modeling, procedural generation, simulation, style transfer and image augmentation [4, 5, 9–11, 18]. These approaches help overcome the challenges of obtaining real-world data, enabling the development of robust and effective computer vision models across various domains [11, 18]. However, these methods of synthetic data generation do not support simulation of physical wear and tear such as rust, primarily because it is effort intensive to create variations in 3D models.

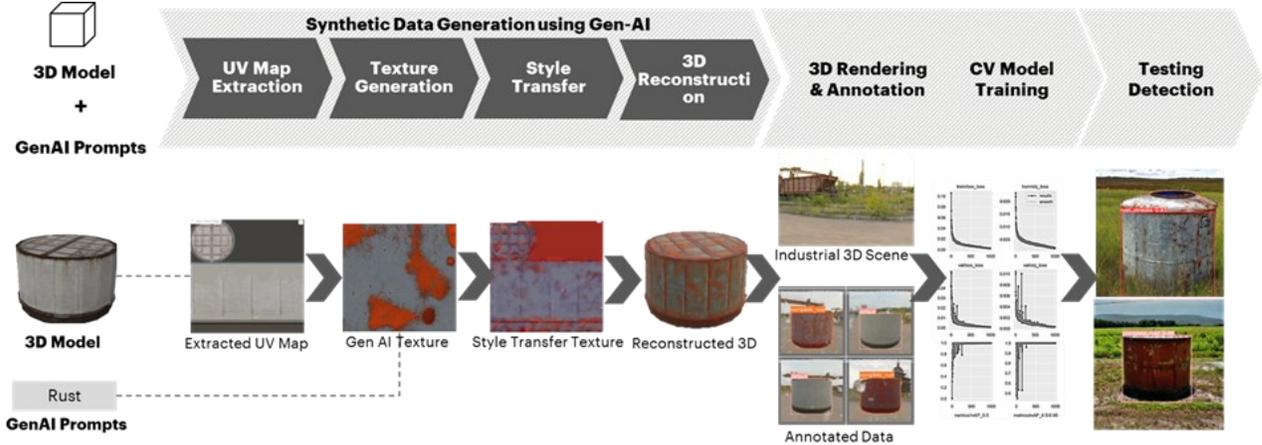

Figure 2: SynGen-Vision: Step-by-step pipeline execution

In recent years, we have seen an emergence of advanced generative models which are pre-trained on extensive datasets, that offer image generation capabilities through a natural language prompt [12, 16, 17, 19]. They generate high-quality images that are virtually indistinguishable from real data. However, these models still occasionally produce images with watermarks, random text, and noisy elements, making them unsuitable for synthetic data generation for industrial applications. Our approach addresses these limitations. It leverages the existing capabilities of foundational models to generate more realistic textures for 3D models. We propose a guided texture generation approach that incorporates style transfer and further noise removal.

Overall, our approach can be differentiated from existing solutions along the following lines. We develop an end-to-end pipeline that generates photorealistic industrial training data. The approach uses GenAI tools along with 3D rendering. Moreover, as our pipeline is generic enough; it can be easily customized for other scenarios with minimal effort.

## 3 Approach

Our proposed GenAI pipeline takes a 3D model of an industrial object, a 3D industrial scene, and a user prompt for a defined usecase, and generates different textures (as shown in Figure 2). Our approach consists of the following steps:

- Texture generation from user prompts
- Texture synthesis with style transfer
- Intelligent texture application on 3D model and 3D scene generation
- Generation of annotated synthetic images from 3D rendering

### 3.1 Texture generation from user prompts

We use stable-diffusion model [16] for text to image generation.

The stable-diffusion model employs a diffusion process to incrementally improve the generated image. This process entails iteratively adjusting the pixel values of an initial image to enhance its quality and match the desired output as specified by the input text.

To generate textures for rust, we experiment with multiple input texts or prompts. Some of the key findings are as follows. Prompts like 'rusted tank', 'rusted can', 'rust spots on can', generate images with background noise, leading to noisy textures. Appending keywords like 'texture' or 'surface' generates cleaner texture images. Keywords like 'complete', 'full' can be used to generate textures with rust all-over. Keywords like 'streaks', 'lines', 'spots', 'slight' can be used to generate textures with less rust. For the given use-case, we use input texts 'complete rust' and 'rust streaks'.

Directly applying the generated textures on the 3D models causes loss of details like patterns, symbols, text, etc. present in the original object texture. Thus, we perform additional steps to retain these details and improve the quality and realism of generated outcomes.

### 3.2 Texture synthesis with style transfer

We use a style-transfer algorithm [9] to synthesize the stylized texture using the base texture of the 3D model and the generated texture from GenAI. The style transfer algorithm takes a content image and a style image and generates a stylized image. This algorithm extracts features from both the images and generates the output by combining the features. This approach retains the details of the original texture, thus creating more realistic and authentic outputs (as shown in Figure 3).

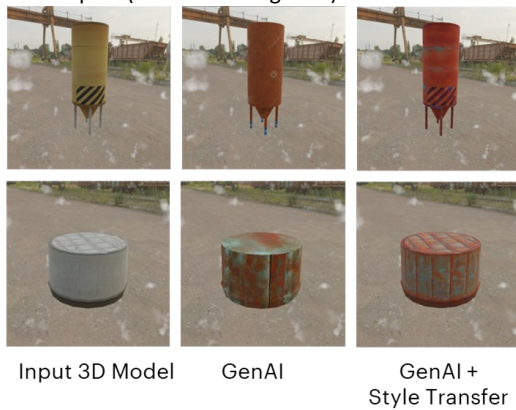

Figure 3: Comparison of outputs generated by GenAI vs. GenAI + Style Transfer approach

### 3.3 Intelligent texture application on 3D models and 3D scene generation

The GenAI approach may generate some unusable textures like those containing text, objects, watermarks or not adhering to the expected degree of rust (as shown in Figure 4). We use image processing techniques [2, 8] to filter out noisy textures. This helps to select the best textures to be applied on the 3D models for rendering. We apply the selected generated textures on the 3D models by updating their UV maps. In the context of 3D graphics, a UV map is a representation of the surface of a 3D model in a 2D space. UV mapping involves unwrapping the 3D model's surface and putting it in a 2D space so that the texture can be applied precisely. We use the 3D rendering tool blender [3] for UV mapping and for placing the modified 3D models in the 3D scene.

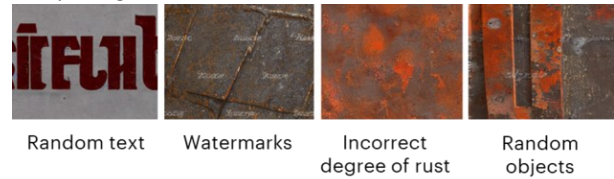

Figure 4: Samples of noisy texture generation

### 3.4 Generation of annotated synthetic images from 3D rendering

We generate multiple variations of the scene by changing camera angles, viewpoints, lighting conditions, etc. We render multiple images and annotate bounding boxes on the industrial object using blender [3]. We save an annotation file containing the image path and the user prompt based class (as shown in Figure 5).

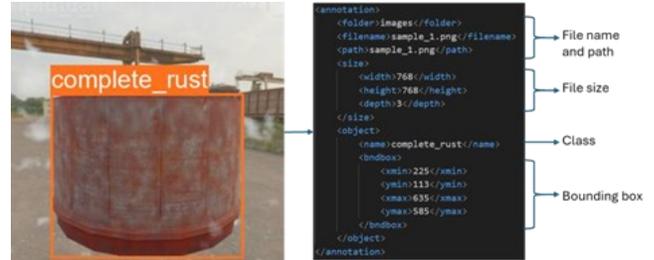

Figure 5: Sample training data

### 3.5 Evaluation and results

To evaluate our approach, we train a computer vision model for rust detection for three classes 1) complete rust 2) rust streaks, and 3) default (no rust). We take 2000 training samples and use yolov5 model [7] for training the rust detection model. The yolov5 model is one of the state-of-the-art solutions for object detection. Each training sample contains an image and a corresponding annotation file. All the training images are synthetically generated using our pipeline.

We evaluate the trained computer vision model on real images. We manually tag about 100 test images for evaluation. The results show a promising outcome for our pipeline.

To the best of our knowledge, we could not find any standard benchmarking dataset for industrial rust detection. We conduct experiments on textures generated using three approaches: a)

GenAI, b) GenAI followed by Style Transfer and c) GenAI followed by Style Transfer and Noise Removal. We perform quantitative evaluation along three metrics – Precision, Recall and mAP50 (as shown in Table 1). mAP50 is a widely used evaluation metric for object detection models and is calculated as the mean average precision for bounding-box overlap with an IoU threshold of 0.5.

pipeline. We compare our solution with multiple approaches and perform significantly better on all the metrics.

With improvements in GenAI approaches, more realistic and refined textures can be generated. The object detection model can also be extended to localize the wear and tear along with classification. SynGen-Vision can be used to expand the scope of

Table 1: Quantitative Analysis

| Approach | Metrics | default (no rust) | rust streaks | complete rust | all |
|---|---|---|---|---|---|
| a) GenAI | Precision | 0.500 | 0.120 | 0.309 | 0.31 |
| | Recall | 0.167 | 0.500 | 0.093 | 0.253 |
| | mAP50 | 0.374 | 0.279 | 0.183 | 0.279 |
| b) GenAI + Style Transfer (ST) | Precision | 1 | 0.250 | 0.250 | 0.583 |
| | Recall | 0.167 | 0.500 | 0.100 | 0.256 |
| | mAP50 | 0.583 | 0.495 | 0.275 | 0.451 |
| c) GenAI + ST + Noise Removal | Precision | 1 | 1 | 1 | 1 |
| | Recall | 0.9 | 0.667 | 0.667 | 0.744 |
| | mAP50 | 0.950 | 0.833 | 0.833 | 0.872 |

Approach (b) performs better than approach (a) as it maintains realism by retaining patterns, symbols and other fine details present in the original texture. Our proposed approach (c) eliminates textures which do not adhere to the expected degree of rust or contain noisy elements like text or watermarks. The trained model performs well on real industrial images with mAP50 value of 0.87. The computer vision model works well even on tanks of different shapes (as shown in Figure 6).

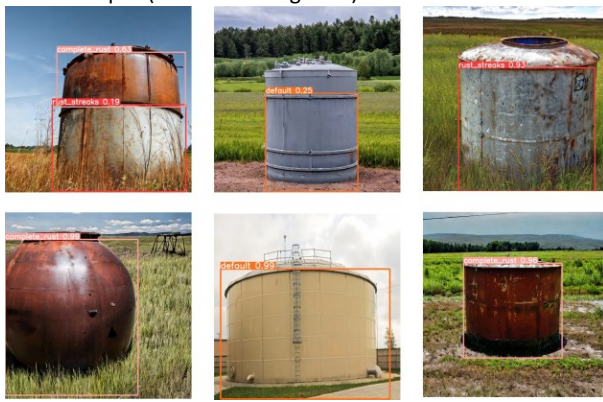

Figure 6: Sample predictions on test data

## 4 Conclusion and Future work

In this work, we propose a pipeline SynGen-Vision, that uses the power of GenAI to synthetically generate and annotate different degrees of rust on industrial objects, for training computer vision models for rust detection. We train the model on a completely generated dataset and test the model on real industrial images. We perform quantitative analysis to evaluate the efficacy of our

Synthetic Data Generation beyond object detection, to simulate various stages of ageing, wear and tear analysis and predictive maintenance. This pipeline aims to improve the process of industrial wear and tear detection with minimal cost and effort.

## Acknowledgments

We would like to thank Elias Treis, Pallavi Ramicetty and Nikita Naik from Accenture Technology Innovation group for the 3D rendering and object detection training pipeline. We would also like to thank Rafael De Souza from Accenture Industry X group, for his guidance, support and industry insights.

## References

[1] Mohammed W Ashour, MM Abdulrazzaq, and Mohammed Siddique. 2023. Machine Vision Inspection of Steel Surface Using Combined Global and Local Features. In *International Conference on Information Technology-New Generations*. Springer, 359–368.

[2] Muhammad Aqeel Aslam, Muhammad Asif Munir, and Daxiang Cui. 2020. Noise removal from medical images using hybrid filters of technique. In *Journal of Physics: Conference Series*, Vol. 1518. IOP Publishing, 012061.

[3] Blender. 2024. Blender. (2024). https://docs.blender.org/api/current/info_overview.html [Online; accessed 31-July-2024].

[4] Cuong Do. 2020. 3D image augmentation using neural style transfer and generative adversarial networks. In *Applications of Digital Image Processing XLIII*, Vol. 11510. SPIE, 707–718.

[5] Shuai Feng, Shengtong Yang, Zhaodong Niu, Jianbin Xie, Mingshan Wei, and Peiqin Li. 2021. Grid cut and mix: flexible and efficient data augmentation. In *Twelfth International Conference on Graphics and Image Processing (ICGIP 2020)*, Vol. 11720. SPIE, 656–662.

[6] Yi Huang. 2023. Intelligent Machine Vision for Detection of Steel Surface Defects with Deep Learning. In *2023 IEEE International Conference on Smart Internet of Things (SmartIoT)*. IEEE, 326–327.

[7] Glenn Jocher, Ayush Chaurasia, Alex Stoken, Jirka Borovec, Yonghye Kwon, Kalen Michael, Jiacong Fang, Colin Wong, Zeng Yifu, Diego Montes, et al. 2022. ultralytics/yolov5: v6. 2-yolov5 classification models, apple m1, reproducibility, clearml and deci. ai integrations. *Zenodo* (2022).

[8] Sukhjinder Kaur. 2015. Noise types and various removal techniques. *International*

*Journal of Advanced Research in Electronics and Communication Engineering (IJARECE)* 4, 2 (2015), 226–230.

[9] Dianqi Li, Yizhe Zhang, Zhe Gan, Yu Cheng, Chris Brockett, Ming-Ting Sun, and Bill Dolan. 2019. Domain Adaptive Text Style Transfer. arXiv:1908.09395 [cs.CL] https://arxiv.org/abs/1908.09395

[10] Hao Li, Xiaopeng Zhang, Qi Tian, and Hongkai Xiong. 2020. Attribute mix: Semantic data augmentation for fine grained recognition. In *2020 IEEE International Conference on Visual Communications and Image Processing (VCIP)*. IEEE, 243–246.

[11] Alhassan Mumuni, Fuseini Mumuni, and Nana Kobina Gerrar. 2024. A survey of synthetic data augmentation methods in computer vision. *arXiv preprint arXiv:2403.10075* (2024).

[12] Taesung Park, Ming-Yu Liu, Ting-Chun Wang, and Jun-Yan Zhu. 2019. Semantic image synthesis with spatially-adaptive normalization. In *Proceedings of the IEEE/CVF conference on computer vision and pattern recognition*. 2337–2346.

[13] Peng Peng and Jiugen Wang. 2019. Wear particle classification considering particle overlapping. *Wear* 422-423 (2019), 119–127. https://doi.org/10.1016/j.wear.2019.01.060

[14] Luca Petricca, Tomas Moss, Gonzalo Figueroa, Stian Broen, et al. 2016. Corrosion detection using AI: a comparison of standard computer vision techniques and deep learning model. In *Proceedings of the sixth international conference on computer science, engineering and information technology*, Vol. 91. AIRCC Publishing Corporation Chennai, India, p99.

[15] Ruosen Qi, Jie Zhang, and Katy Spencer. 2023. A Review on Data-Driven Condition Monitoring of Industrial Equipment. *Algorithms* 16, 1 (2023). https://doi.org/10.3390/a16010009

[16] Robin Rombach, Andreas Blattmann, Dominik Lorenz, Patrick Esser, and Björn Ommer. 2022. High-resolution image synthesis with latent diffusion models. In *Proceedings of the IEEE/CVF Conference on Computer Vision and Pattern Recognition*. 10684–10695.

[17] Chitwan Saharia, William Chan, Saurabh Saxena, Lala Li, Jay Whang, Emily L Denton, Kamyar Ghasemipour, Raphael Gontijo Lopes, Burcu Karagol Ayan, Tim Salimans, et al. 2022. Photorealistic text-to-image diffusion models with deep language understanding. *Advances in neural information processing systems* 35 (2022), 36479–36494.

[18] Frederik Seiler, Verena Eichinger, and Ira Effenberger. 2024. Synthetic Data Generation for AI-based Machine Vision Applications. *Electronic Imaging* 36, 6 (2024), 276–1–276–1. https://doi.org/10.2352/EI.2024.36.6.IRIACV-276

[19] Gabriele Valvano, Antonino Agostino, Giovanni De Magistris, Antonino Graziano, and Giacomo Veneri. 2024. Controllable Image Synthesis of Industrial Data using Stable Diffusion. In *Proceedings of the IEEE/CVF Winter Conference on Applications of Computer Vision*. 5354–5363.

[20] Miao Wang, Lei Yang, Zhibin Zhao, and Yanjie Guo. 2022. Intelligent prediction of wear location and mechanism using image identification based on improved Faster R-CNN model. *Tribology International* 169 (2022), 107466. https://doi.org/10.1016/j.triboint.2022.107466